\documentclass[letterpaper, 10 pt, conference]{ieeeconf}  

\IEEEoverridecommandlockouts                              

\usepackage{graphicx} 
\usepackage{wrapfig}
\usepackage{hyperref}
\usepackage{color}
\usepackage{tabu}
\usepackage{amsmath}
\usepackage{rotating}
\usepackage{multirow}
\usepackage[table]{xcolor}    
\usepackage{booktabs}

\makeatletter
\DeclareRobustCommand\onedot{\futurelet\@let@token\@onedot}
\def\@onedot{\ifx\@let@token.\else.\null\fi\xspace}

\makeatother

\definecolor{gred}{RGB}{101,115,40}
\definecolor{red1}{RGB}{255,102,102}
\definecolor{green1}{RGB}{102,178,102}
\definecolor{yellow1}{RGB}{255,128,16}
\definecolor{blue1}{RGB}{48,130,137}
\definecolor{halfred}{RGB}{128,0,0}

\title{\LARGE \bf Failure Prediction for Autonomous Driving}

\author{Simon Hecker$^{1}$, Dengxin Dai$^{1}$, and Luc Van Gool$^{1,2}$
\thanks{$^{1}$Simon Hecker, Dengxin Dai, and Luc Van Gool are with the Toyota TRACE-Zurich team at the Computer Vision Lab, 
        ETH Zurich, 8092 Zurich, Switzerland
        {\tt\small firstname.lastname@vision.ee.ethz.ch }}%
\thanks{$^{2}$Luc Van Gool is also with the Toyota TRACE-Leuven team at the Dept of Electrical Engineering ESAT, KU Leuven
         3001 Leuven, Belgium}%
}

\begin{document}

\maketitle
\thispagestyle{empty}
\pagestyle{empty}

\begin{abstract}
The primary focus of autonomous driving research is to improve driving accuracy. While great progress has been made, state-of-the-art algorithms still fail at times. Such failures may have catastrophic consequences. It therefore is important that automated cars foresee problems ahead as early as possible. This is also of paramount importance if the driver will be asked to take over. We conjecture that failures do not occur randomly. For instance, driving models may fail more likely at places with heavy traffic, at complex intersections, and/or under adverse weather/illumination conditions. This work presents a method to learn to predict the occurrence of these failures, i.e. to assess how difficult a scene is to a given driving model and to possibly give the human driver an early headsup. A camera-based driving model is developed and trained over real driving datasets. The discrepancies between the model's predictions and the human `ground-truth' maneuvers were then recorded, to yield the `failure' scores. Experimental results show that the failure score can indeed be learned and predicted. Thus, our prediction method is able to improve the overall safety of an automated driving model by alerting the human driver timely, leading to better human-vehicle collaborative driving. 
\end{abstract}

\section{INTRODUCTION}
\label{sec:introduction} 



Autonomous vehicles will have a substantial impact on people's daily life, both personally and professionally. For instance, automated vehicles can largely increase human productivity by turning driving time into working time, provide personalized mobility to non-drivers, reduce traffic accidents, or free up parking space and generalize valet service~\cite{autonomous:vehicle:guide:policymakers}. As such, developing automated vehicles is becoming the core interest of many, diverse industrial players. Recent years have witnessed great progress in autonomous driving \cite{deep:driving,nvidia:driving:16,end:driving:16,perception:path:generation,chen2017brain,arxiv2018:hecker}, resulting in announcements that autonomous vehicles have driven over many thousands of miles and that companies aspire to sell such vehicles in a few years. All this has fueled expectations that fully automated vehicles are coming soon. 

Yet, significant technical obstacles must be overcome before assisted driving can be turned into full-fletched automated driving, a prerequisite for the above visions to materialize. To make matters worse, an automated car that from time to time will call on the driver to take over, will, by many drivers, be considered worse than having no automated driving at all. Indeed, in such a transition situation, the driver will be required to permanently pay attention to the road, as to not be out of context when s/he suddenly needs to act. And that does not go together well with the boredom coming with not having to intervene for a long time. The more successful the automation, the worse the issue. Add legal responsibilities to the picture, and the possibility that the human driver is called upon to take decisions, however rarely that is, may still be with us for a while. 

\begin{figure*}
\centering
\includegraphics[width=0.85\linewidth]{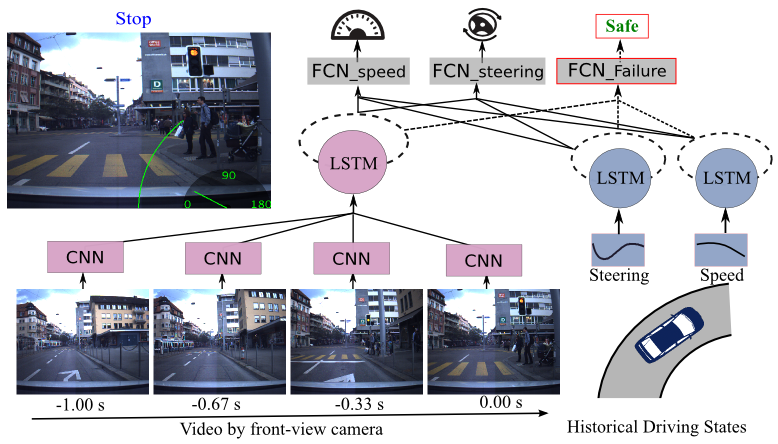}
\caption{The architecture of our driving model which provides future maneuvers (i.e. speed and steering angle) and the drivability score of the scene. The drivability scores are quantized into two levels: \emph{Safe} and \emph{Hazardous}. In this case, the coming scene is \emph{Safe} for the driving model, so the system does not need to alert the human driver. FCN stands for fully connected network.}
\label{fig:pipeline}
\end{figure*}

With so much effort currently going into improving autonomous driving, such systems will certainly improve quickly. Yet, as said, during the coming years performance will probably not be strong enough such that occasional mistakes can be avoided altogether. Indeed, driving models may still fail due to congested traffic, bad weather, frontal illumination, road constructions, etc., or simply unexpectedly, due to the idiosyncrasies of the underlying algorithms. Failures of a vehicle can be catastrophic~\cite{the:cost:of:accident}, and it is therefore crucial to obtain an early warning for impending trouble. 
Despite this importance, the community has so far paid limited attention to the automated predictions of potential failures. We therefore decided to push for a capability where driving models can yield a warning such as \emph{I am unable to make a reliable decision for the coming situation}, and can give the human driver an early warning about a possible need for human intervention. 


We propose the concept of \emph{Scene Drivability}, i.e. how easy a scene is for an automated car to navigate. A low \emph{drivability} score means that the automated vehicle is likely to fail for the particular scene. Obviously, \emph{scene drivability} is dependent on the autonomous driving system at hand. In order to quantify and learn this property, we therefore first need to pick a particular autonomous driving model. We developed one of our own, solely based on video observations. Videos from car-mounted cameras were used to train it. 
In keeping with modern machine learning, it automatically learned things like `when the vehicle is in the left-most lane, the only safe maneuvers are a right-lane change or keeping straight, unless the vehicle is approaching an intersection'. It is clear that such learning requires the system to be exposed to a representative sample of scenarios. We therefore trained the model on a large, real driving dataset, which contains video sequences and other time-stamped sensor measurements such as steering angles, speeds, and GPS coordinates \cite{arxiv2018:hecker}. The driving model achieves a performance similar to other recent approaches based on video observations~ \cite{end:driving:16,driving:attention,arxiv2018:hecker}. Discrepancies between the predictions by the trained driving model and the ground-truth maneuvers by human drivers are then used to assess the likelihood of failure, i.e. the \emph{Scene drivability} score.  

Due to the success of deep neural networks in supervised learning \cite{lecun2015deep} and especially in autonomous driving, we develop a Recurrent Convolutional Network (RCNet) with four CNNs~\cite{vgg16} as visual encoders and three LSTMs~\cite{lstm} to integrate the visual contents, temporal relationships, and the previous driving states (steering angle and speed) into one single prediction model. The model can be trained very efficiently in an end-to-end manner, and its architecture is shown in Figure~\ref{fig:pipeline}. This architecture is used for both  tasks: car driving and its failure prediction. All layers, except for the task-specific fully-connected layers, are shared for computational efficiency.

Readers will notice that our model is quite simple. The emphasis of this paper is not on achieving the state-of-the-art driving performance. Rather, it is to provide a sensible driving model and infer failure prediction for it, as a contribution to let autonomous driving survive the risky market situation ahead. The choice of the model is also due to our access to sensors and data. 

In this work, we quantize the \emph{scene drivability} scores for particular driving scenes to two levels: \emph{Safe} and \emph{Hazardous}. They are intended to translate to the two driving modes `Full Automation' and `Driver assisted'. Our experiments show that scene drivability can indeed be learned and predicted. 
Of course, the drivability will increase if the driving model is improved, especially when information from other sensors is added, such as from GPS, laser scanners, and radar. Our method is flexible enough to include those. We also do {\em not} claim to predict the drivability of scenes for any driving model out there, but rather propose a framework that can be trained to extract the drivability for other models as well.

\section{Related Work} 
\label{sec:related}


Our work is relevant to both autonomous and assisted driving, and to vision system failure mode prediction. 

\subsection{Driving Models for Automated Cars}

Significant progress has been made in autonomous driving, especially due to the deployment of deep neural networks. Driving models can be clustered into two groups based on their working paradigms~\cite{deep:driving}: mediated perception approaches and end-to-end mapping approaches.

Mediated perception approaches require recognition of all driving-relevant objects, such as lanes, traffic signs, traffic lights, cars, pedestrians, etc.~\cite{kitti:dataset,Cityscapes,3d:object:detection:AD}. Some of these recognition tasks could be tackled separately, and there are excellent works~\cite{3d:traffic:scene} integrating the results. This group of methods represents the current state-of-the-art for autonomous driving, and most of the results are reported with diverse sensors used, such as laser scanners, GPS, radar and high-definition maps of the environment. 


End-to-end mapping methods aim to construct a direct mapping from the sensory input to the maneuvers. The idea can be traced back to the 1980s, when a neural network was used to learn a direct mapping from images to steering angles~\cite{network:autonomous:1980}. Another successful example of learning a direct mapping is~\cite{LeCun:driving:05}, which uses ConvNets to learn a human driver's steering angles. The popularity of this idea is fueled by the success of end-to-end trained deep neural networks and the availability of large driving datasets~\cite{nvidia:driving:16,oxford:driving,arxiv2018:hecker}. 
Recent advances have been shown in \cite{arxiv2018:hecker} by incorporating surround-view videos and route planning. The future end-to-end approaches may also need a  mixture of sensors and modules for even better performance. Possible modules consist of traffic agent detection and tracking~\cite{target:tracking:lidar:13,multi-sensor:detection:tracking:14,3d:object:detection:AD,semantic:foggy:scene,DomainAdaptiveFasterRCNN}, future prediction of road agents' location and behavior ~\cite{predicting:deeper,sceneparse:opticalflow,intent:aware:pedestrian:prediction}, and driveability map generation~\cite{driveability:maps:for:lane}.


\noindent
\subsection{Assistive Features for Vehicles}
Over the last decades, more and more assistive technologies have been deployed, that help to increase driving safety.  Technologies such as lane keeping, blind spot checking, forward collision avoidance, adaptive cruise control etc., alert drivers about potential dangers~\cite{forecast:control, control:driver:modeling, kasper2012object}. In the same vein, drivers are monitored to avoid distraction and drowsiness~\cite{look:at:driver, gaze:driver}, and maneuvers are anticipated~\cite{car:knows:iccv15} to generate alerts in a timely manner. Readers are referred to \cite{looking:at:human} for an excellent overview of such work. Our work complements existing ADAS and driver monitoring techniques by equipping fully automated cars with an assistive feature to anticipate automation failures and yields a timely alert for the human driver to take over. 


%
%


\subsection{Failure Prediction}
Performance-blind algorithms can be disastrous. As automated vision increasingly penetrates industrial applications, this issue is gaining attention~\cite{dai:phd:thesis,zhang2014predicting}. Notable examples in computer vision for learning model uncertainty or failure include: semantic image segmentation \cite{kendall2015bayesian}, optical flow \cite{confidence:of:08, flow:confidence}, image completion \cite{completion:quality}, stereo \cite{stereo:confidence}, and image creation \cite{dai:synthesizability}. Our work adds autonomous driving to the list. In addition to creating warnings, performance-aware algorithms bring other benefits as well. For instance, they can speed up algorithms downstream, by adaptively allocating computing resources based on scene difficulty. For autonomous driving, this can also mean using sensors adaptively or selectively. Another paper relevant to ours is \cite{Anticipating:accidents:DB:18}, which anticipates traffic accidents by learning from a large-scale incidents database. 

\section{Method} 
\label{sec:method}
In this section, we first present our end-to-end direct mapping method for autonomous driving, based on the recent success of recurrent neural networks. We then present how we use the same architecture to learn to predict the (un)certainty of the system, i.e. our drivability score. 

\begin{figure*}
\centering
$\begin{tabular}{cc}
\includegraphics[width=0.45\linewidth]{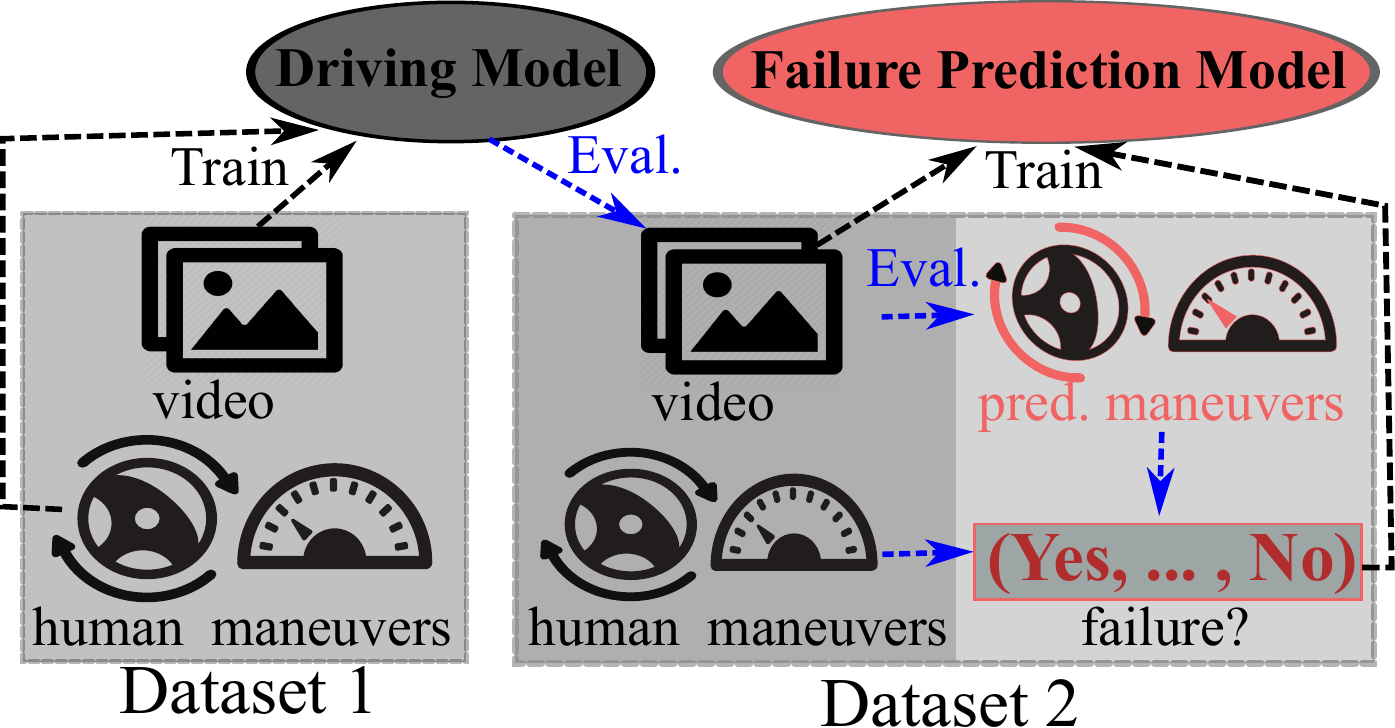} & 
\includegraphics[width=0.38\linewidth]{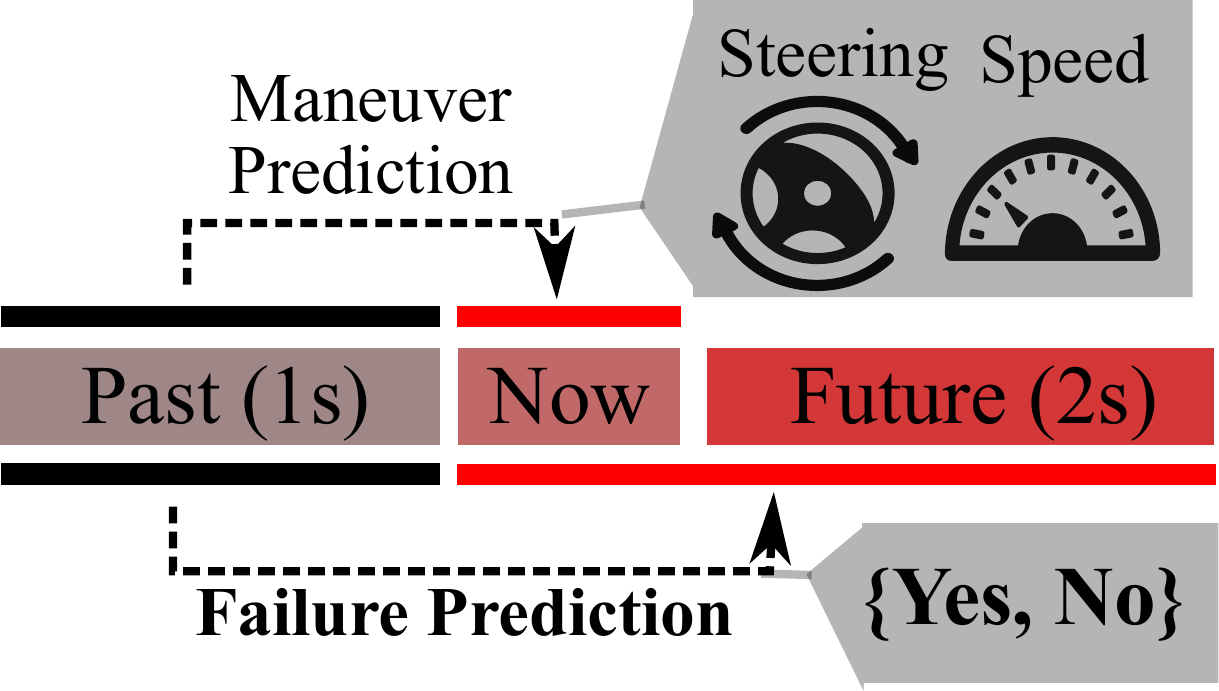} \\
\text{(a) training procedure}  & 
\text{(b) target space}
\end{tabular}$
\caption{An illustrative flowchart of the training procedure and solution space of our driving model and the failure prediction model.}
\label{fig:timeline}
\end{figure*}

\subsection{Driving Model}
\label{sec:drivingmodel}

In contrast to predicting the car's ego-motion like previous work~\cite{nvidia:driving:16,end:driving:16}, our model predicts the steering wheel (angle) and the speed of the cars directly. The goal of our driving model is to map directly from a frontal-view video to the steering wheel angle and speed of the car. Let us denote the video by $V$, and the vehicle's steering wheel angle and speed by $A$ and $S$ respectively. We assume that the driving model works with discrete time and makes driving decisions every $1/f$ seconds. The inputs $V$, $A$ and $S$ are synchronized and sampled at a sampling rate $f$. In this work, $f=4$. Unless stated otherwise, our inputs and outputs all are represented in this discretized form.   

Let us denote the current video frame by $V_t$, the current vehicle's speed by $S_t$, and the current steering angle by $A_t$. The $k^{th}$ previous values can then be denoted by $V_{(t-k)}$, $S_{t-k}$ and $A_{t-k}$, resp., and all $k$ previous values can be denoted by $\mathbf{V}_{[t-k,t)} \equiv \langle V_{t-k}, ..., V_{t-1}\rangle$, $\mathbf{A}_{[t-k,t)} \equiv \langle A_{t-k}, ..., A_{t-1}\rangle$ and $\mathbf{S}_{[t-k,t)} \equiv \langle S_{t-k}, ..., S_{t-1}\rangle$, resp.. 
Our goal is to train a deep network that predicts driving actions from all inputs: 
\begin{equation}
F: (\mathcal{V}_{[t-k,t]}, \mathcal{A}_{[t-k,t-1)}, \mathcal{S}_{[t-k,t-1)}) \rightarrow \mathcal{A}_t \times \mathcal{S}_t
\label{eq:learning:fun1}
\end{equation}
where $\mathcal{A}_t$ represents the steering angle space and $\mathcal{S}_t$ the speed space for the current time.  
$\mathcal{A}$ and $\mathcal{S}$ can be defined at several levels of granularity. We consider the continuous values directly recorded from the car's CAN bus, where $\mathcal{S}=\{S | 0  \leq S \leq 180$ for speed and $\mathcal{A}=\{A | -720  \leq A \leq 720\}$ for steering angle. Here, kilometer per hour (km/h) is the unit of $S$, and degree ($^\circ$) the unit of $A$. 

Given $N$ training samples collected during real drives, learning to predict the driving actions for the current situation is based on minimizing the following cost: 
\begin{equation}
\begin{split}
L(\theta) = \sum_{n=1}^N \Big( l({A}^{n}_t, F_{\text{a}}(\mathbf{V}^{n}_{[t-k, t]}, \mathbf{A}^{n}_{[t-k, t)}, \mathbf{S}^{n}_{[t-k, t)})) \\ + \lambda  l({S}^{n}_t, F_{\text{s}} (\mathbf{V}_{[t-k, t]}^n, \mathbf{A}^{n}_{[t-k, t)}, \mathbf{S}^{n}_{[t-k, t)})) \Big), 
\end{split}
\end{equation}
where $\lambda$ is a parameter balancing the two losses, one for steering angle and the other for speed. We use $\lambda=1$ in this work due to prior CAN signal normalization. $F$ is the learned function for the driving model. For the continuous regression task, $l(.)$ is the $L2$ loss function. Our model learns from multiple previous frames in order to better understand traffic dynamics. We assume that the \emph{current} video frame $V_t$ is already available for making the decision. 

\subsection{Failure Prediction}

An automated car can fail due to many causes. Here we focus on \emph{scene drivability} -- a driving situation is too challenging for the driving model to make reliable decisions. We define failure scores based on the discrepancies between the predicted maneuvers (steering angles and speed) and the human driver's maneuvers. In particular, we denote the predicted speed and steering angle by $\bar{S}_t$ and $\bar{A}_t$. Then, the failure for speed and steering angle estimation are signaled by: 
\begin{equation}
g^a_t = \text{sgn}(\|A_t-\bar{A_t}\|-T_a), 
\end{equation}
and 
\begin{equation}
g^s_t = \text{sgn}(\|S_t-\bar{S_t}\|-T_s), 
\end{equation}
where 
\begin{equation}
\text{sgn}(x) =
\begin{cases}
1 & \text{if } x \ge 0,\\
0 & \text{if } x < 0.
\end{cases}, 
\end{equation} 
and $T_a$ and $T_s$ are thresholds defining correct and incorrect predictions for steering angle and speed. Then, the failure occurrence for the current time is signaled by: 
\begin{equation}
g_t = g_t^a \vee g_t^s, 
\label{eq:failure:1}
\end{equation}
where $\vee$ is an OR operator: $x \vee y = 0$ if $x = y = 0$ and $x \vee y = 1$ otherwise.

The definition by Equation \ref{eq:failure:1} quantizes scene drivability  into two levels: \emph{Safe} and \emph{Hazardous}. \emph{Safe} scenes are defined as those with an absolute error of less than $T_{a}$ degrees for steering angle and with an absolute error of less than $T_{s}$ km/h for speed. \emph{Hazardous} are those with a deviation in either category larger than the defined threshold. A \emph{safe} scene allows for a driving mode of \emph{High Automation} and a \emph{hazardous} scene allows for a driving mode of \emph{Partial/No Automation}. These thresholds can be set and tuned according to specific driving models and legal regulations. 
 

Failure prediction is more useful, the earlier it can be done, i.e. the more time can be given to the human driver to take over. Therefore, the failure that our model is trained to predict is from current time $t$ to future time $t+m$: 
\begin{equation} 
g_{\overrightarrow{[t,t+m]}} = g_t \vee g_{t+1} \vee g_{t+2} \vee ... \vee g_{t+m}.
\label{eq:failure:2}
\end{equation}
By learning to predict $g_{\overrightarrow{[t,t+m]}}$, our model will alert the human driver if either the speed prediction and/or the steering angle prediction is going to fail at any of the time points in the time period $[t, t+m]$.  
The learning goal is then changed to training a deep network model to make a prediction for driving actions for current time $t$ and to make a prediction for the drivability score for the time period from $t$ to a future time point $t+m$. In particular, the learning target is changed from Equation~\ref {eq:learning:fun1} to: 
\begin{equation}
F: (\mathcal{V}_{[t-k,t]}, \mathcal{A}_{[t-k,t-1)}, \mathcal{S}_{[t-k,t-1)}) \rightarrow \mathcal{A}_t \times \mathcal{S}_t \times \mathcal{G}_{\overrightarrow{[t,t+m]}}
\label{eq:learning:fun2}
\end{equation}
where $\mathcal{G}_{\overrightarrow{[t,t+m]}}=\{0,1\}$ denotes the space of our drivability score defined by Equation~\ref{eq:failure:2}. In this work, $m$ is set to $8$ to represent a period of $2$ seconds. A different length can be used if the application needs. Please see Figure~\ref{fig:timeline} for the illustrative flowchart of the training procedure and solution space of our driving model and the failure prediction model. 


\subsection{Implementations}
\label{sec:implementation}
We adopt a deep neural network for our learning task. The model learns to predict three targets: the vehicle's steering angle, its speed, and the failure score defined by Equation~\ref{eq:failure:2}. In particular, our model consists of four copies of convolutional neural networks (CNNs) with shared weights as visual encoders, combined with three Long Short Term Memory networks (LSTMs) to integrate the visual information, historical driving speed, and historical steering angles. The outputs of the three LSTMs are integrated by three fully connected networks (FCN) to make the final predictions for the vehicle's steering angle, its speed, and the failure status of Equation~\ref{eq:failure:2}. As shown in Figure~\ref{fig:pipeline}, all layers of the network are shared by the three tasks except for the top, task-specific layers. 

\begin{figure*}
  \begin{center}
    \includegraphics[width=0.75\textwidth]{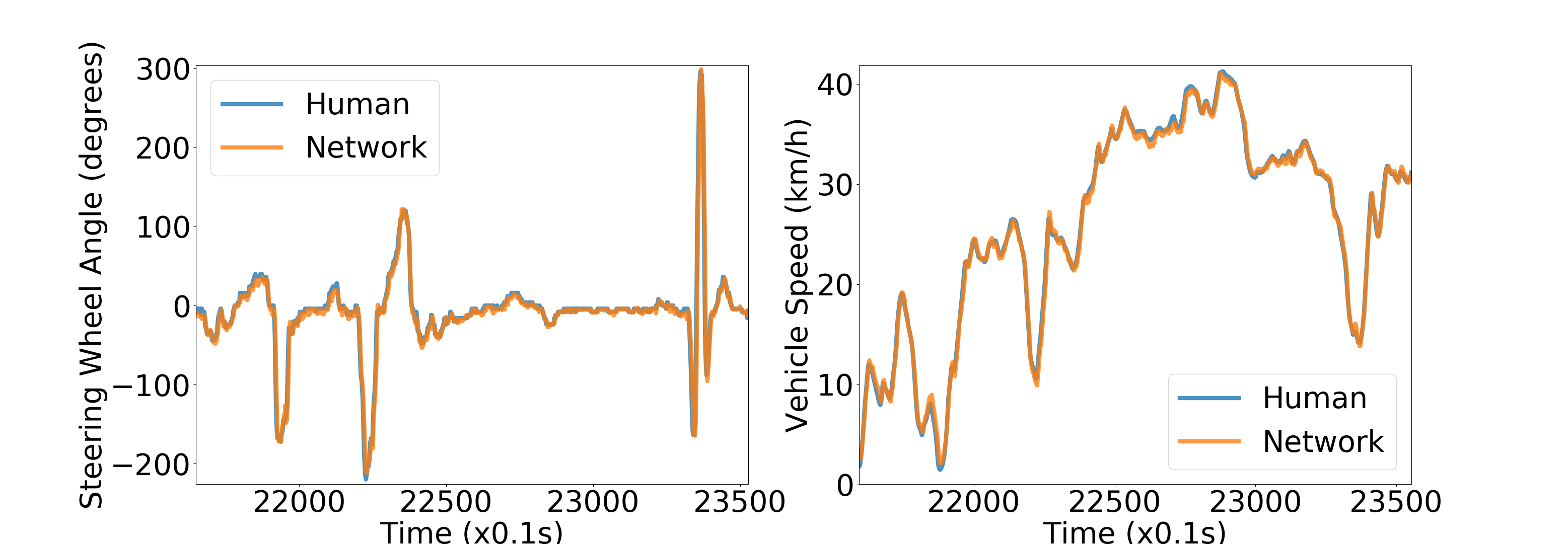}
  \end{center}
  \caption{Visualization of driving model performance for steering wheel angle and vehicle speed prediction. The overlap areas signal the consistency between the estimations by our driving model (Network) and the measurements from human drivers (Human). }
  \label{fig:performance:plot} 
\end{figure*}

The image input sequence consists of four video frames, taken from a front-facing camera mounted on the roof of the vehicle. The sequence includes the current video frame along with three previous frames; this allows for images in the sequence to vary significantly and thus improves the predictive performance of the model. Input images are center cropped to $270\times270$ pixels from an initial resolution of $480\times270$, resized to $240\times240$ and finally randomly cropped to $224\times224$ during training and center cropped to the same dimensions during evaluation. Each image in the sequence is fed into a ResNet34~\cite{resnet} with shared convolutional layers. The convolutional layers are  pre-trained on the classification task of ImageNet~\cite{imagenet}. All layers of the ResNet are trainable with the final layers output being fed into a 2-layer FCN: $fc(1024)$ - Relu - $fc(1024)$ - Relu. The parameters of the FCN are randomly initialized.  
This results in a $4\times1024$ feature vector which describes the high level historical and current visual input into the system.

We incorporate three parallel LSTMs~\cite{lstm} with $128$, $16$ and $16$ hidden states and $4$, $2$ and $2$ layers, resp. The high level visual features of the ResNets, the historical speed information, and the steering angle information are fed into the three LSTMs, resp. Steering angle and speed information are sampled at the same sampling rate of $1/f=4$, the same as for the video. A temporal sequence of length $k=4$ is used for all three inputs. 

At this point, we have aggregated a total of three feature vectors, that describe visual information, historical steering angle and historical speed. These vectors are then concatenated.  Our final prediction task varies depending on whether we train a driving agent or a failure prediction agent. Consequently the very top layers of our model architecture will vary depending on the task. In the driving agent case, the network is continuously trained to output the current steering angle and vehicle speed using two regression networks. The two regression networks consist of a 2-layer fully connected network of $fc(512)$ - Relu - $fc(1)$ each, and are tasked to output either steering angle or vehicle speed. For the failure prediction agent, we train our network on a two class \textit{Safe} or \textit{Hazardous} classification task as defined by Equation~\ref{eq:failure:2}. We predict to an interval of $m=8$ samples (i.e. $2$ seconds) into the future from the current time. This gives us the opportunity to notify the driver of a potential hazard ahead of time, allowing for an adequate response. Our classification task network architecture consists of a 2-layer fully connected network $fc(512)$ - Relu - $fc(2)$. This network is optimized via the cross entropy loss. 

We optimize our network with the Adam Optimizer~\cite{adam} and a learning rate of $0.00001$. Our models train for 10 epochs with a mini-batch size of $32$ which results in around $15$ hours of training time each with a GeForce GTX TITAN X Graphics Card. 

\begin{figure*}
$\begin{tabular}{cccc}
\begin{turn}{90}\hspace{0mm}  $[T_a, T_s] = [10.0^{\circ}, 5.0\text{ km/h}]$ \hspace{2mm} $[T_a, T_s] = [7.0^{\circ}, 3.0\text{ km/h}]$  \hspace{4mm} $[T_a, T_s] = [5.0^{\circ}, 2.0\text{ km/h}]$  \end{turn} &
\hspace{-1mm}
\includegraphics[width=0.95\linewidth, height=131mm]{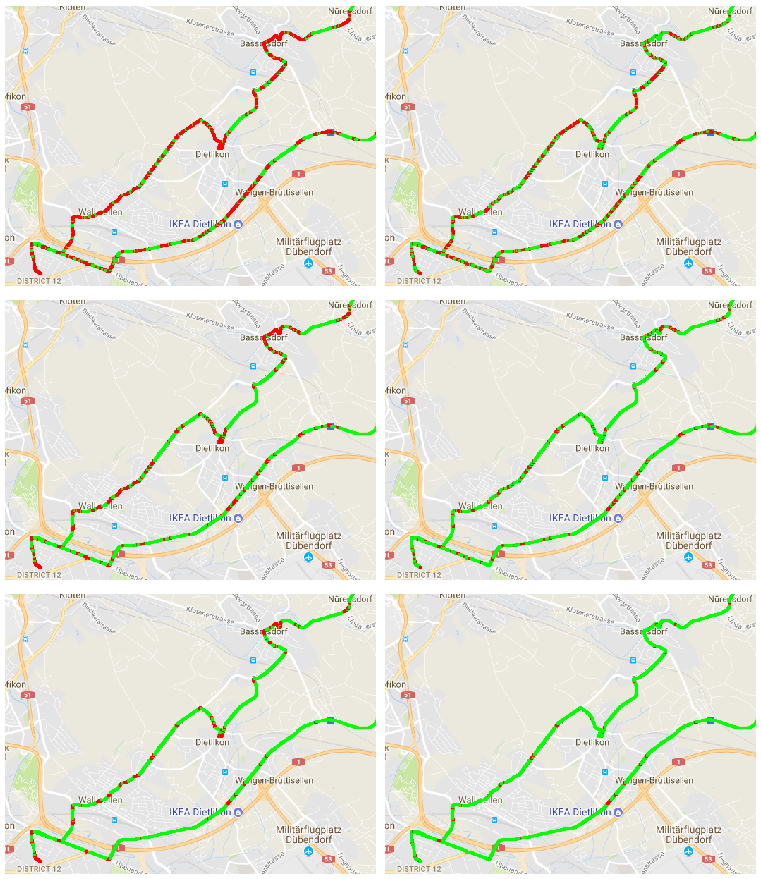}\\ 
& \hspace{15mm} (a) Driving Model \hspace{35mm} (b) Driving Model + Failure Prediction Model
\end{tabular}$

\caption{Visualization of our driving model performance in the left column and our failure prediction model improvement, when allowing 25\% manual driving time, in the right column. Failure is quantized to either: $[T_a, T_s] = \{[5.0^{\circ}, 2.0\text{ km/h}], [7.0^{\circ}, 3.0\text{ km/h}], [10.0^{\circ}, 5.0\text{ km/h}]\}$ for each row respectively. Scenes are classified as being either \emph{Safe} ( \fcolorbox{black}{green}{\rule{0pt}{3pt}\rule{3pt}{0pt}} ) or \emph{Hazardous} ( \fcolorbox{black}{red}{\rule{0pt}{3pt}\rule{3pt}{0pt}} ). As an example, \emph{Safe} is defined as scenes for which the deviation of the predicted steering angle and vehicle speed to the human ground-truth is smaller than the respective threshold; \emph{Hazardous} are those with a deviation in either category larger than the respective threshold. The right column shows the reduction of \emph{Hazardous} scenes, for each threshold, due to 25\% \emph{human driving}. Thus reclassifying the top 25\% most hazardous scenes, identified by the network, as \emph{Safe}.}


\label{fig:failures} 
\end{figure*}



\section{Experiments}
\label{sec:experiment}

\subsection{Datasets and Training}
We train and evaluate our method on our autonomous vehicle dataset which consists of around $150,000$ unique sequences captured by a car mounted camera in Switzerland. Alongside the video data, time-stamped sensor measurements are provided by the dataset as well, such as the vehicle's speed, steering wheel angle and GPS locations. Thus, this data is ideal for self-driving studies. The GPS coordinates allow for compelling visualizations of where the model fails. In order to properly train and evaluate our model, we split our dataset to three datasets of equal size: Dataset 1, Dataset 2, and Dataset 3. We train our driving model on Dataset 1, train the failure prediction model on Dataset 2, and evaluate both models on Dataset 3. The two models need to be trained on separate datasets, because the predictions of the driving model on its own training set are too optimistic to reflect the real failures. Please refer to Figure~\ref{fig:timeline} for the training procedure of our models.


\begin{figure*}[!tb]
$\begin{tabular}{cccc}
\hspace{-2mm}
\includegraphics[width=0.33\linewidth]{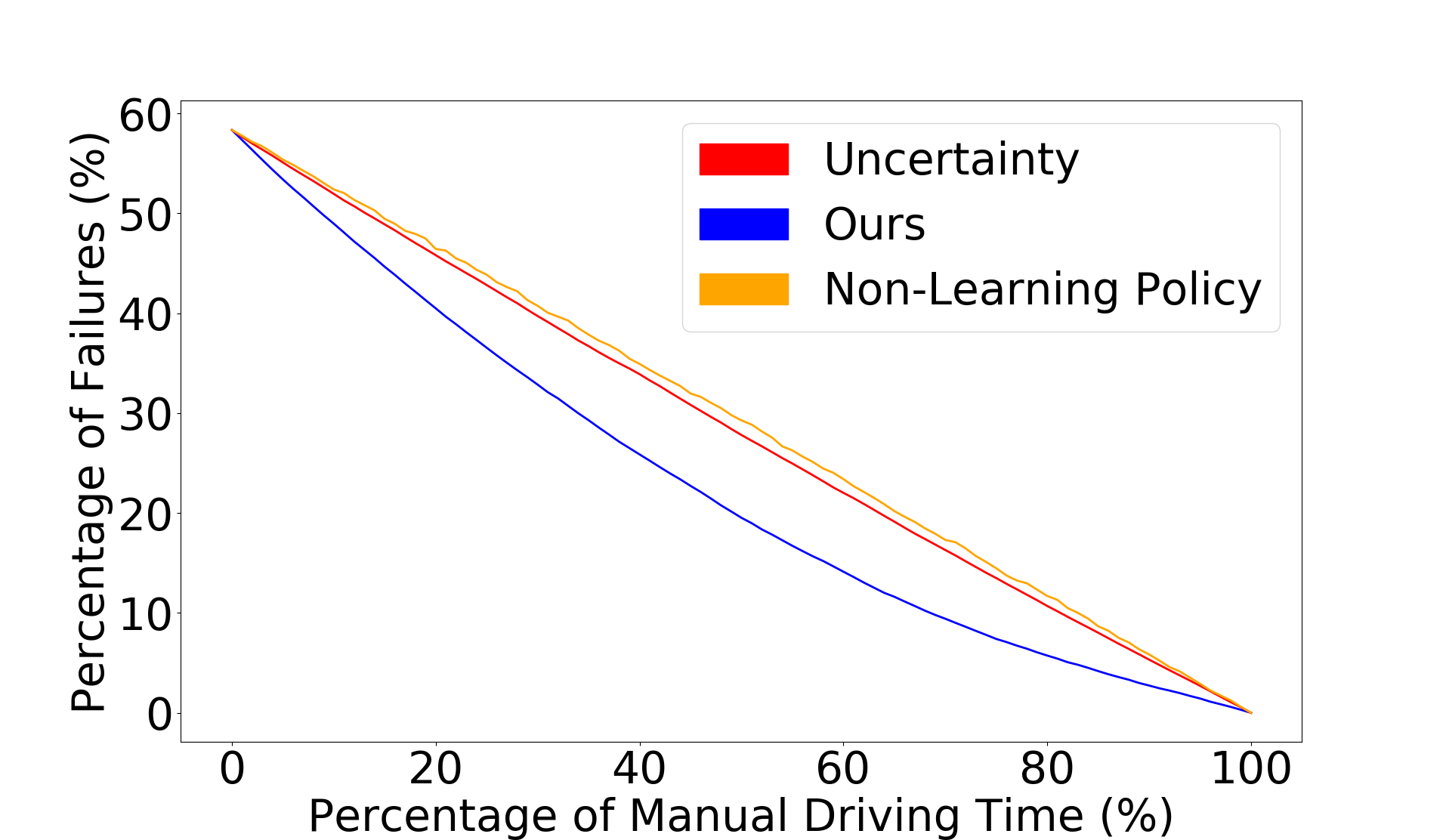}
& \hspace{-4mm}
\includegraphics[width=0.33\linewidth]{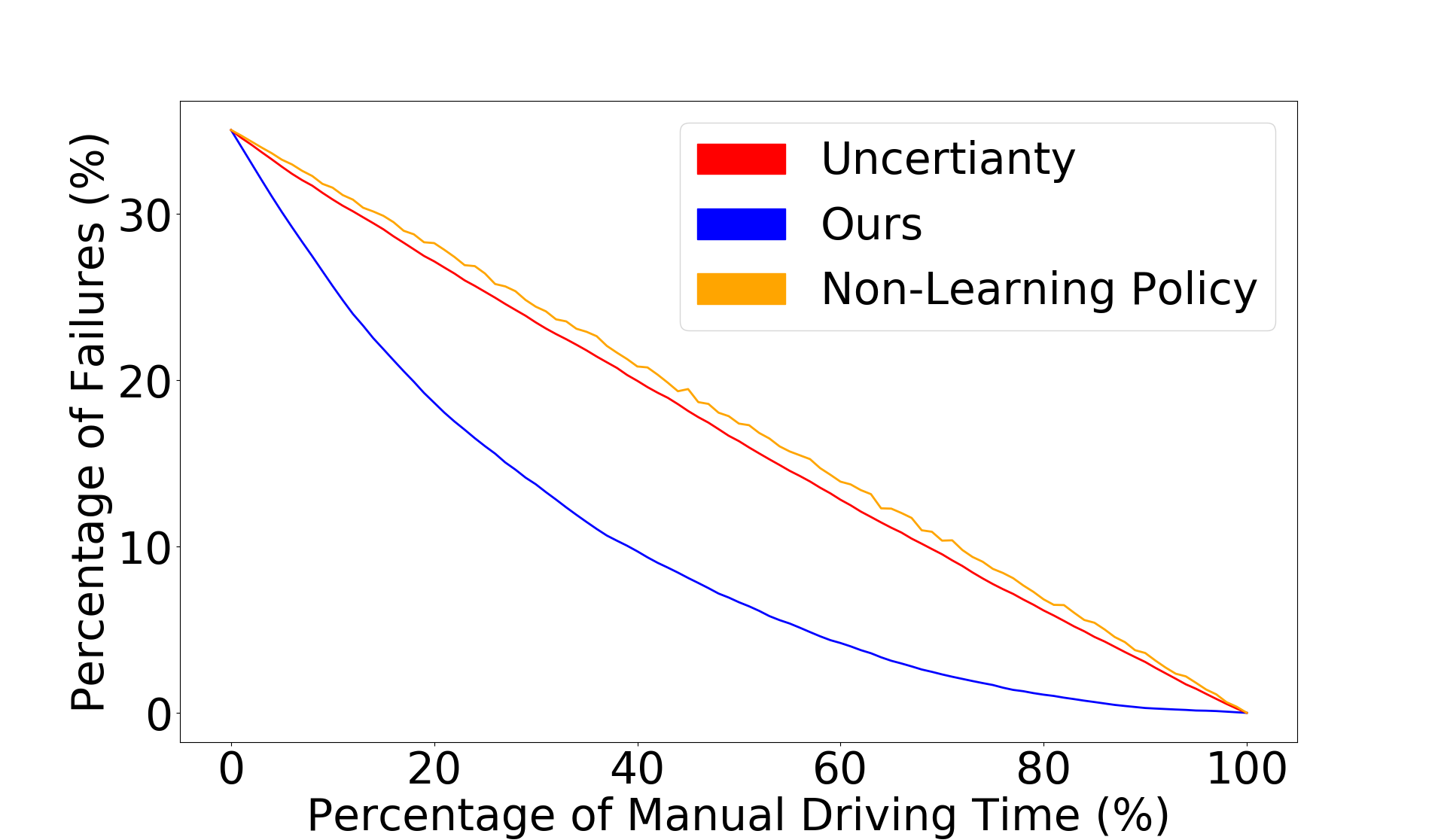}
& \hspace{-4mm}
\includegraphics[width=0.33\linewidth]{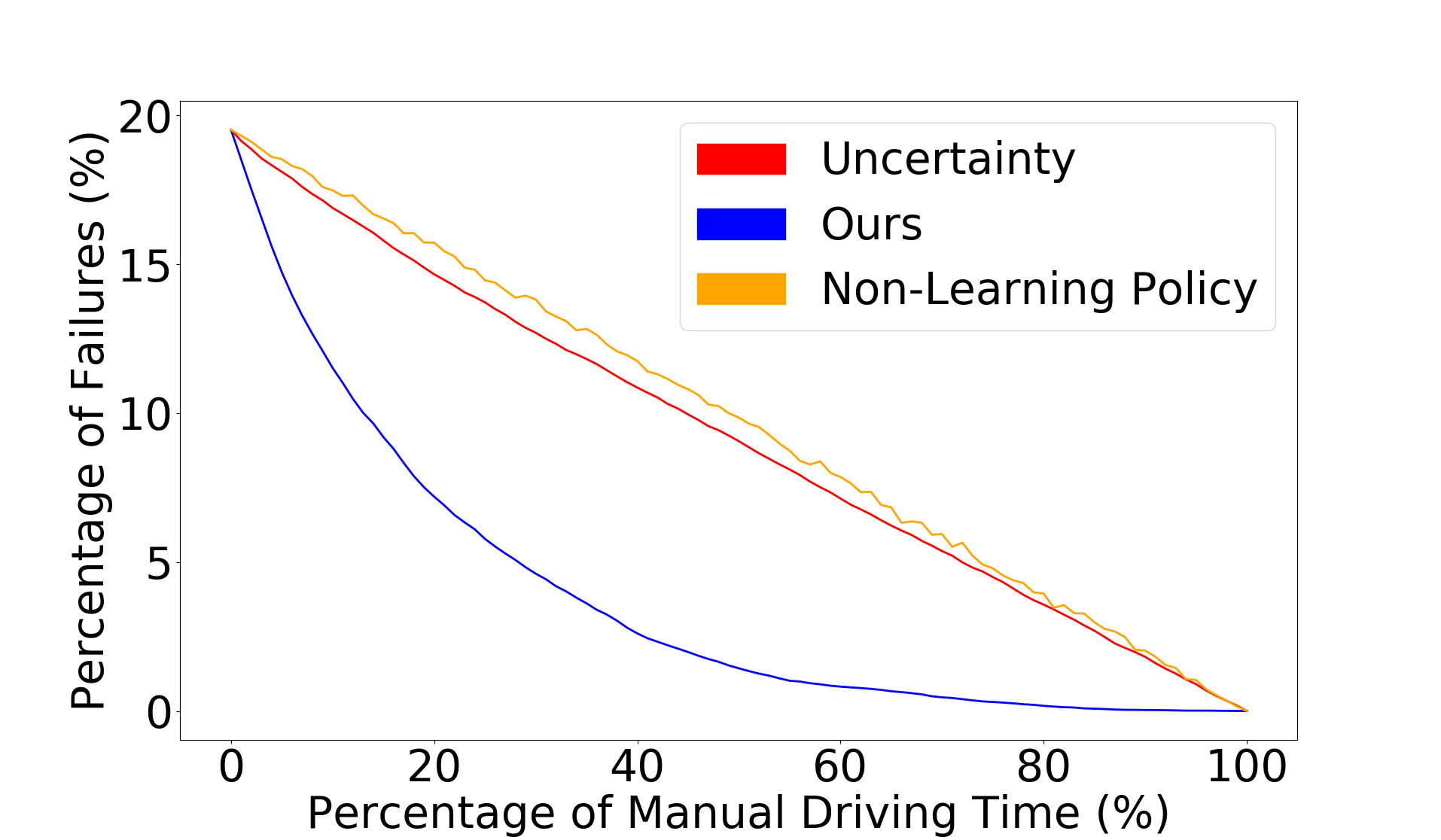} \\   
\text{(a) $[T_\theta, T_v] = [5.0^{\circ}, 2.0\text{ km/h}]$ } &  \text{(b) $[T_\theta, T_v] = [7.0^{\circ}, 3.0\text{ km/h}]$} & \text{(c) $[T_\theta, T_v] = [10.0^{\circ}, 5.0\text{ km/h}]$ } \\
\end{tabular}$
\caption{The reduction of failures (\%) as a function of the percentage of human driving time (\%). In this hybrid, semi-automatic driving system, human drivers are alerted of potential failures and are asked to take over. Our failure prediction model is compared to a basic model which does not have a learning policy and alerts the driver at regular intervals, and to an uncertainty based approach~\cite{dropout:bayesian}. 
}  
\label{fig:failurePrediction} 
\end{figure*}

\begin{table}[!tb]
  \centering
  \caption[Control Performance]{Control Performance. Comparison between mean absolute error (MAE) in m/s and degree. }
  \label{tab:stateoftheart}
    \begin{tabular}{| l | l | l |}
    \hline
     \textbf{Model} & \textbf{MAE speed} &\textbf{MAE angle}  \\ \hline
  	CNN+LSTM \cite{driving:attention} & N.A. & 4.15   \\ \hline
    Our Model & 0.15 & 3.66 \\ \hline
    \end{tabular} \vspace{-4mm}
\end{table}

\subsection{Driving Accuracy}
We first compare the overall performance of our driving model to the state-of-the-art models based on video observations \cite{end:driving:16,driving:attention}, and find that our model yields results similar to these models. This is not only illustrated in Figure~\ref{fig:performance:plot}, where our driving model is very close to predicting the human ground truth driving performance, but also in Table \ref{tab:stateoftheart} that highlights the control performance in terms of mean absolute error. 
\begin{figure*}[!tb]
\centering
$\begin{tabular}{cccc}
\begin{turn}{90}\hspace{1cm} \emph{Safe}\end{turn} &
\hspace{-4mm}
\includegraphics[width=0.31\linewidth]{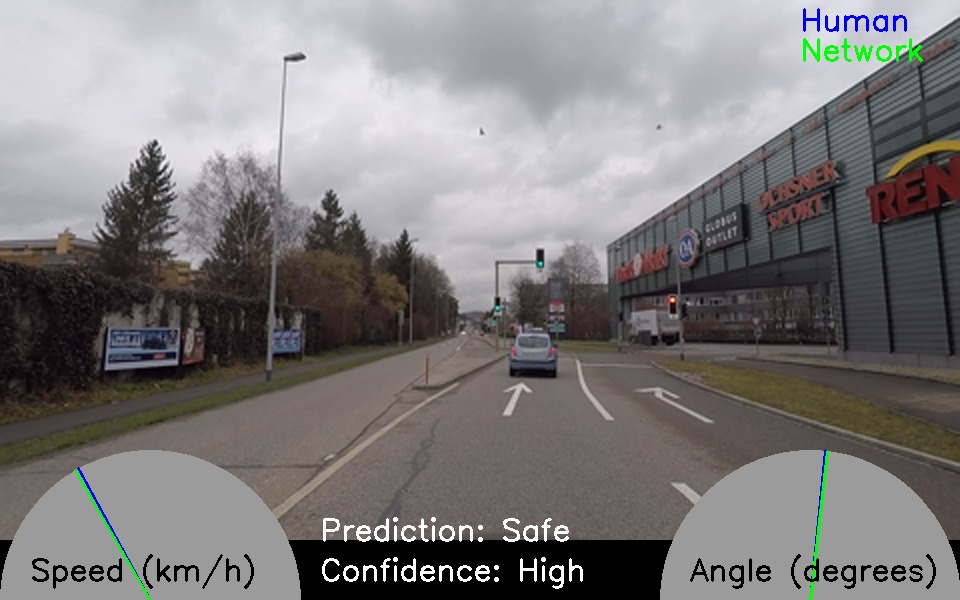}
& \hspace{-4mm}
\includegraphics[width=0.31\linewidth]{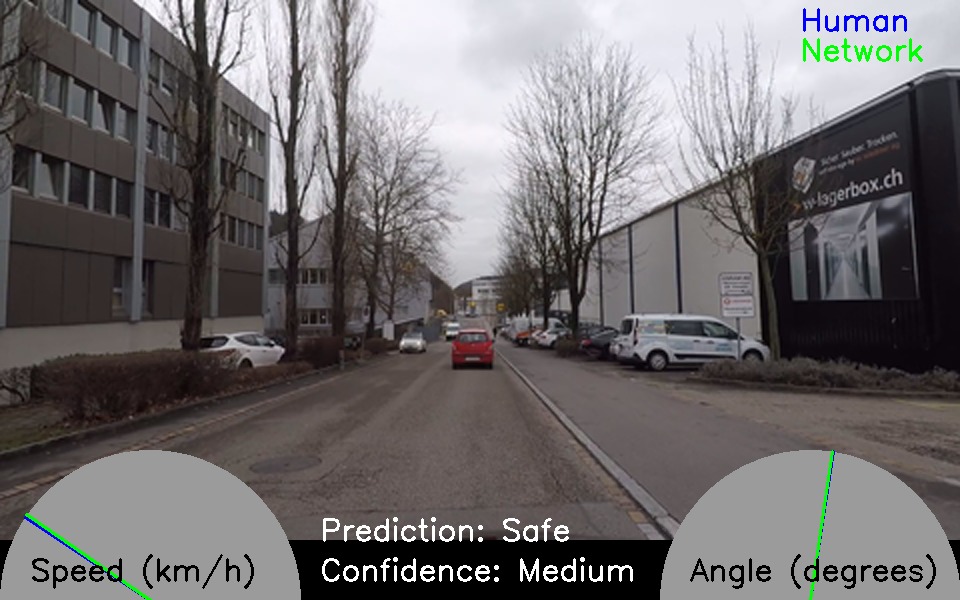}
& \hspace{-4mm}
\includegraphics[width=0.31\linewidth]{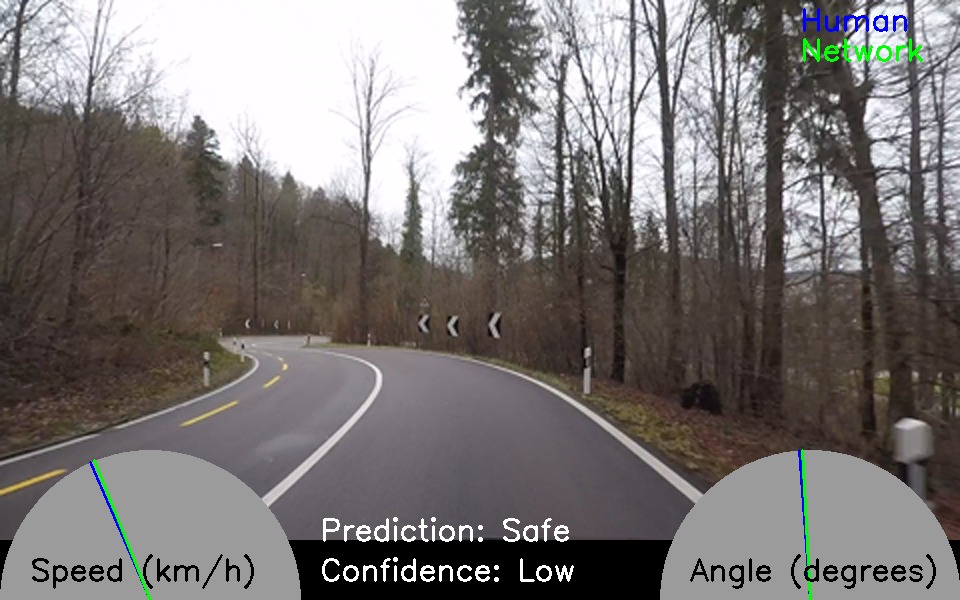} \\ 
\begin{turn}{90}\hspace{1cm} \emph{Hazardous}\end{turn} &
\hspace{-4mm}
\includegraphics[width=0.31\linewidth]{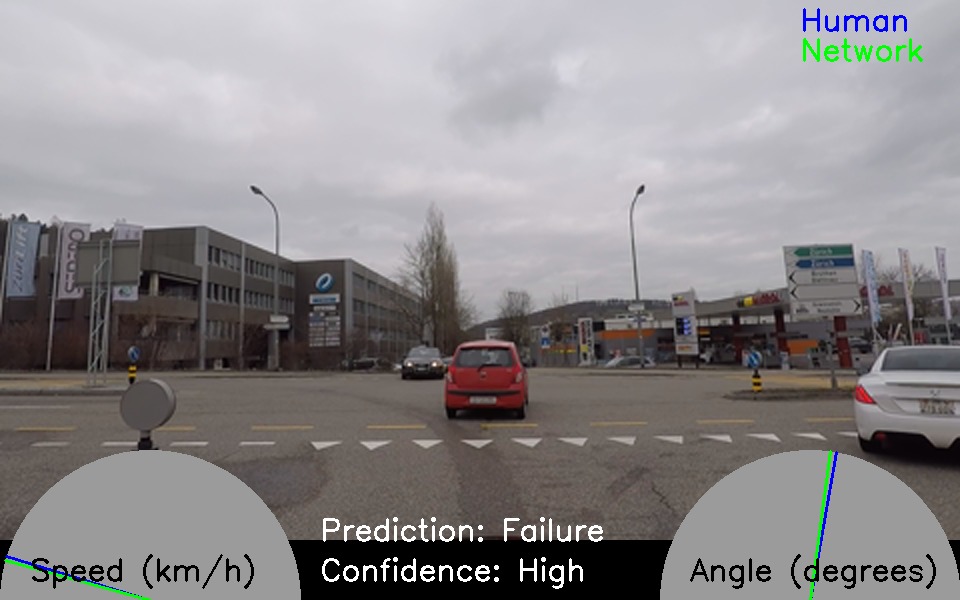}
& \hspace{-4mm}
\includegraphics[width=0.31\linewidth]{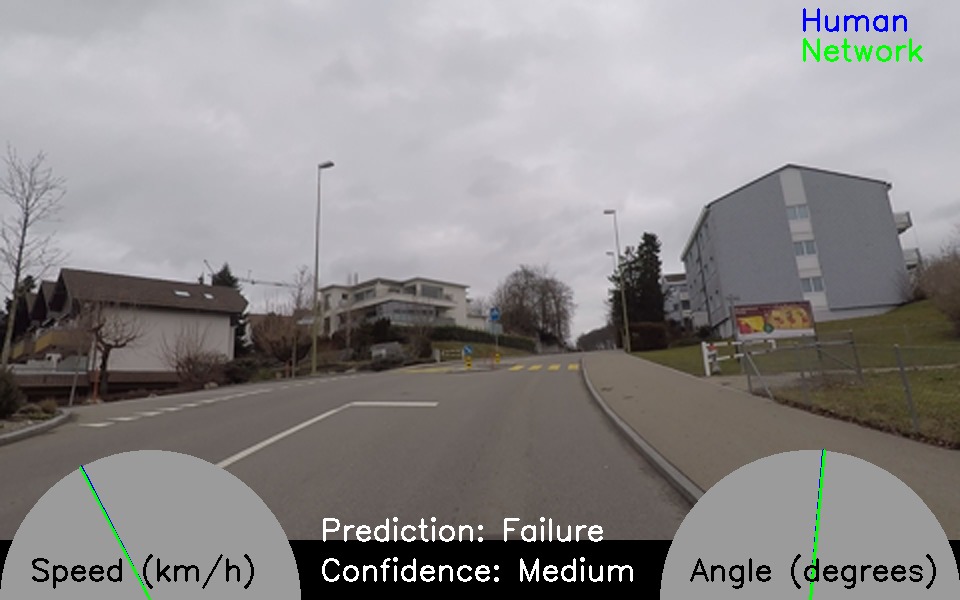}
& \hspace{-4mm}
\includegraphics[width=0.31\linewidth]{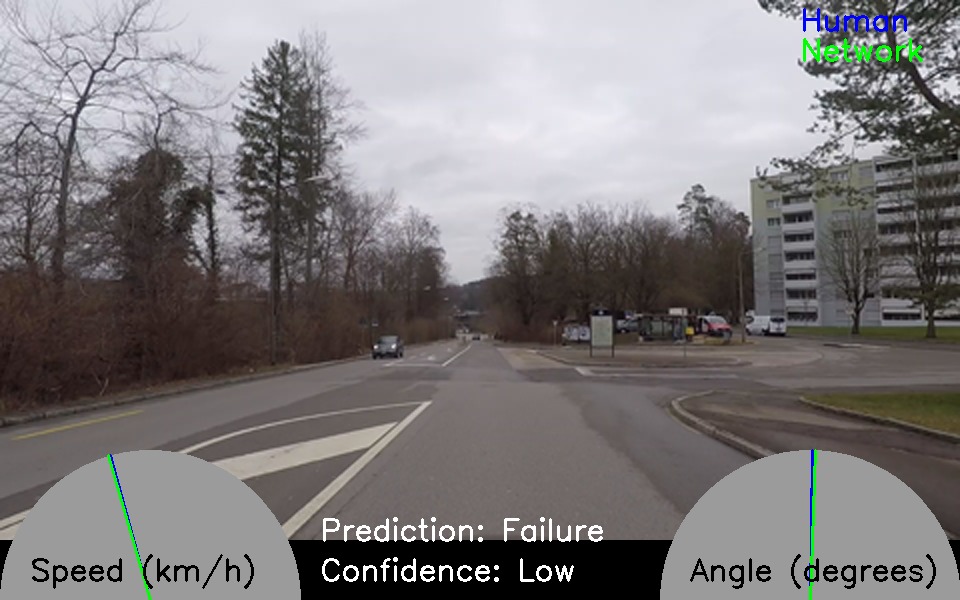} \\  
 & \text{ High} &  \text{ Medium} & \text{ Low } \\
\end{tabular}$
\caption{Scene examples with their drivability scores, quantized into two levels: \emph{Safe} and \emph{Hazardous} for rows, and the confidence of the prediction in columns ranging from high to low. The steering angle and driving speed by human drivers ( \fcolorbox{black}{blue}{\rule{0pt}{3pt}\rule{3pt}{0pt}} ) and by our driving model ( \fcolorbox{black}{green}{\rule{0pt}{3pt}\rule{3pt}{0pt}} ) are shown as well. Note that failures are predicted for 2 seconds in the future, whereas driving decisions are only for the current time.  
}  
\label{fig:examples:pics} 
\end{figure*}

In addition to reporting the overall quantitative numbers, we visualize where the most common failures of our driving model occur. This is achieved by defining a failure of the driving model when either the predicted vehicle speed or steering wheel angle deviates more than a defined threshold, and plotting the performance as a function of color on the map. We show, in Figure~\ref{fig:failures}, three different threshold settings: $[T_s, T_a] \leftarrow [2\text{ km/h},5^{\circ}]$, $[T_s, T_a)] \leftarrow [3\text{ km/h}, 7^{\circ}]$,  and $[T_s, T_a] \leftarrow [5\text{ km/h}, 10^{\circ}]$, ranging from a maximum deviation of 5 degrees and 2 km/h for our tightest definition of failure up to 10 degrees and 5 km/h for our loosest definition of failure. 

In particular, the model is more likely to fail at intersections, partially due to  the unknown destination of the vehicle and thus the ambiguity of which route the driver will take. We acknowledge that this is largely due to the lack of route planning in our driving model. However, route planning has been used to improve driving models in our recent work \cite{arxiv2018:hecker} and a fusion of route planned aware driving models with failure prediction is considered as our next future work. In addition, we mainly observed failures during sharp corners, in congested traffic, and in urban environments when many pedestrians are involved. 


\subsection{Failure Prediction}

In this section, we evaluate our failure prediction model. 
Accurate and timely requests for manual take-over do result in a safety gain. We now show that our model learns to accurately alert the driver of impeding driving agent failure, reducing the risk of collisions. This is also illustrated by the gain in safety for different budgets of manual driving time ranging from $1-100\%$ of total driving time. 

For this work, we train our failure prediction model on three different thresholds for speed and angle. These range from a very tight definition of failure using a threshold of 5 degrees and 2 km/h, to 7 degrees and 3 km/h, and finally to a loose definition of 10 degrees and 5km/h. We use the same underlying driving agent for each of these three instances, and thus obtain different metrics for our failure prediction dataset. It is worth noticing that three failure prediction models are trained, one for each threshold setting. 

We evaluate the method in the setting of image retrieval. The scenes for which the driving model is mostly likely to fail are retrieved and handed over to the human driver to deal with. We then  compute the reduction of failure as a function of manual driving time.  We compare our method to a baseline method which requires the same amount of manual driving time but does not have a learned policy about when to ask for manual intervention; it rather notifies the driver at regular intervals to take control. 
We also compare our method to an uncertainty estimation method for neural networks~\cite{dropout:bayesian}, for which the human driver is asked to take over when the driving model is mostly uncertain about its output. The uncertainty is computed by the dropout technique~\cite{dropout:bayesian}. 

The results in Figure~\ref{fig:failurePrediction} show that our method can effectively reduce the amount of driving-model induced failure by switching to manual driving timely and accurately. Our model performs significantly better than the baseline model and the uncertainty model~\cite{dropout:bayesian}, because our method is specifically trained for the purpose. This trend can also be observed in Table ~\ref{tab:gainSafety}, where we depict the percentage gained safety over the baseline model for all three threshold models. The experimental results show that failures of a driving model can be learned and predicted quite accurately, and the failure prediction can be used to improve  driving safety in a human-vehicle collaborative driving system. Predicting failure is actually easier than predicting the correct driving decisions, and thus all the more worthwhile including. 

While our method performs better in all three cases, we do notice a more pronounced improvement when our definition of failure is more lax.  One possible reason is that for the case with a very strict definition of the failure, the noise in the recordings of low-level maneuvers is too influential.

\begin{table}[!tb]
  \centering
  \caption[Control Performance]{Safety gain of our failure prediction over a baseline method in a hybrid, human-vehicle collaborative driving system. }
  \setlength{\tabcolsep}{3pt} 
\rowcolors{1}{}{lightgray}    
\begin{tabular}{|c|ccc|} 
\rowcolor{gray!50} 
    &  \multicolumn{3}{c|}{$[T_s, T_a]$}  \\
   Manual Driving & $[5.0^{\circ}, 2 \text{ km/h}]$ & $[7.0^{\circ}, 3 \text{ km/h}]$  &$[10.0^{\circ}, 5 \text{ km/h}]$ \\ 
  	10.0\% & +6.9\% & +23.1\% & +52.5\%   \\
    15.0\% & +11.0\% & +35.5\% & +81.2\%    \\ 
    20.0\% & +15.1\% & +51.2\% & +117.4\%    \\ 
    25.0\% & +19.0\% & +64.7\% & +153.7\%    \\ 
    30.0\% & +23.8\% & +78.7\% & +195.0\%    \\ 
    35.0\% & +29.6\% & +96.4\% & +249.4\%    \\ 
    40.0\% & +34.9\% & +116.2\% & +349.9\%    \\  \hline
    \end{tabular}
    \label{tab:gainSafety}
\end{table}

Finally, we show in Figure~\ref{fig:examples:pics} several driving scenes with the predicted maneuvers (speed and steering angle) and the drivability scores. While the method is able to alert drivers that a driving scene is hazardous for the driving model, it is often hard to figure out the underlying reason. A brief explanation such as `too many road constructions' or `road getting too narrow' will significantly reduce the confusion caused to the driver.  An investigation into such underlying reasons is future work. 



\section{Conclusion}
\label{sec:conclusion} 
In this work, we have presented the concept of \emph{Scene Drivability} for automated cars. It indicates how feasible a particular driving scene is for a particular automated driving method. In order to quantify it, we have developed a novel learning method based on recurrent neural networks. We treated the discrepancies between the predictions of the automated driving model and the human drivers' maneuvers as the (un)drivability scores of the scenes. Experimental results show that such drivability scores can be learned and predicted, and the prediction can be used to improve the safety of automated cars. The learning framework is flexible and can be applied to other driving models with more sensors. To the best of our knowledge, this is the first attempt to predict the failures of automated driving models.
Our future work includes 1) developing more sophisticated driving models (e.g. including recognition of traffic relevant objects, route planning, and 360 degree sensing); 2) extending our failure prediction model to the new driving models; and 3) adding diagnostics by making explicit the inferring reasons for the failures.

\vspace{3mm}
\noindent
\textbf{Acknowledgment}: This work is supported by Toyota Motor Europe via the research project TRACE-Zurich.

%
\bibliographystyle{ieee}
\bibliography{egbib}

\begin{thebibliography}{10}\itemsep=-1pt

\bibitem{autonomous:vehicle:guide:policymakers}
J.~M. Anderson, K.~Nidhi, K.~D. Stanley, P.~Sorensen, C.~Samaras, and O.~A.
  Oluwatola.
\newblock {\em Autonomous vehicle technology: A guide for policymakers}.
\newblock Rand Corporation, 2014.

\bibitem{flow:confidence}
O.~M. Aodha, A.~Humayun, M.~Pollefeys, and G.~J. Brostow.
\newblock Learning a confidence measure for optical flow.
\newblock {\em IEEE Transactions on Pattern Analysis and Machine Intelligence},
  35(5):1107--1120, 2013.

\bibitem{nvidia:driving:16}
M.~Bojarski, D.~Del~Testa, D.~Dworakowski, B.~Firner, B.~Flepp, P.~Goyal, L.~D.
  Jackel, M.~Monfort, U.~Muller, J.~Zhang, et~al.
\newblock End to end learning for self-driving cars.
\newblock {\em arXiv preprint arXiv:1604.07316}, 2016.

\bibitem{the:cost:of:accident}
G.~Calabresi.
\newblock {\em The Cost of Accidents: A Legal and Economic Analysis}.
\newblock Yale University Press, 1970.

\bibitem{perception:path:generation}
L.~Caltagirone, M.~Bellone, L.~Svensson, and M.~Wahde.
\newblock Simultaneous perception and path generation using fully convolutional
  neural networks.
\newblock {\em arXiv:1703.08987}, 2017.

\bibitem{forecast:control}
A.~Carvalho, S.~Lef{\'e}vre, G.~Schildbach, J.~Kong, and F.~Borrelli.
\newblock Automated driving: The role of forecasts and uncertainty—a control
  perspective.
\newblock {\em European Journal of Control}, 24:14--32, 2015.

\bibitem{deep:driving}
C.~Chen, A.~Seff, A.~Kornhauser, and J.~Xiao.
\newblock Deepdriving: Learning affordance for direct perception in autonomous
  driving.
\newblock In {\em International Conference on Computer Vision}, 2015.

\bibitem{chen2017brain}
S.~Chen, S.~Zhang, J.~Shang, B.~Chen, and N.~Zheng.
\newblock Brain inspired cognitive model with attention for self-driving cars.
\newblock {\em arXiv:1702.05596}, 2017.

\bibitem{3d:object:detection:AD}
X.~Chen, H.~Ma, J.~Wan, B.~Li, and T.~Xia.
\newblock Multi-view 3d object detection network for autonomous driving.
\newblock In {\em CVPR}, 2017.

\bibitem{DomainAdaptiveFasterRCNN}
Y.~Chen, W.~Li, C.~Sakaridis, D.~Dai, and L.~{Van Gool}.
\newblock Domain adaptive faster r-cnn for object detection in the wild.
\newblock In {\em IEEE Conference on Computer Vision and Pattern Recognition
  (CVPR)}, 2018.

\bibitem{multi-sensor:detection:tracking:14}
H.~Cho, Y.~W. Seo, B.~V. K.~V. Kumar, and R.~R. Rajkumar.
\newblock A multi-sensor fusion system for moving object detection and tracking
  in urban driving environments.
\newblock In {\em IEEE International Conference on Robotics and Automation
  (ICRA)}, 2014.

\bibitem{target:tracking:lidar:13}
J.~Choi, S.~Ulbrich, B.~Lichte, and M.~Maurer.
\newblock Multi-target tracking using a 3d-lidar sensor for autonomous
  vehicles.
\newblock In {\em International IEEE Conference on Intelligent Transportation
  Systems (ITSC)}, 2013.

\bibitem{Cityscapes}
M.~Cordts, M.~Omran, S.~Ramos, T.~Rehfeld, M.~Enzweiler, R.~Benenson,
  U.~Franke, S.~Roth, and B.~Schiele.
\newblock The cityscapes dataset for semantic urban scene understanding.
\newblock In {\em IEEE Conference on Computer Vision and Pattern Recognition
  (CVPR)}, 2016.

\bibitem{dai:phd:thesis}
D.~Dai.
\newblock {\em Towards Cost-Effective and Performance-Aware Vision Algorithms}.
\newblock PhD thesis, ETH Zurich, 2016.

\bibitem{dai:synthesizability}
D.~Dai, H.~Riemenschneider, and L.~{Van Gool}.
\newblock The synthesizability of texture examples.
\newblock In {\em IEEE Conference on Computer Vision and Pattern Recognition
  (CVPR)}, 2014.

\bibitem{imagenet}
J.~Deng, W.~Dong, R.~Socher, L.-J. Li, K.~Li, and L.~Fei-Fei.
\newblock Imagenet: A large-scale hierarchical image database.
\newblock In {\em IEEE Conference on Computer Vision and Pattern Recognition
  (CVPR)}, 2009.

\bibitem{dropout:bayesian}
Y.~Gal and Z.~Ghahramani.
\newblock Dropout as a bayesian approximation: Representing model uncertainty
  in deep learning.
\newblock In {\em International Conference on Machine Learning (ICML)}, 2016.

\bibitem{3d:traffic:scene}
A.~Geiger, M.~Lauer, C.~Wojek, C.~Stiller, and R.~Urtasun.
\newblock 3d traffic scene understanding from movable platforms.
\newblock {\em IEEE transactions on pattern analysis and machine intelligence},
  36(5):1012--1025, 2014.

\bibitem{kitti:dataset}
A.~Geiger, P.~Lenz, C.~Stiller, and R.~Urtasun.
\newblock Vision meets robotics: The kitti dataset.
\newblock {\em The International Journal of Robotics Research},
  32(11):1231--1237, 2013.

\bibitem{resnet}
K.~He, X.~Zhang, S.~Ren, and J.~Sun.
\newblock Deep residual learning for image recognition.
\newblock In {\em IEEE Conference on Computer Vision and Pattern Recognition
  (CVPR)}, 2016.

\bibitem{arxiv2018:hecker}
S.~{Hecker}, D.~{Dai}, and L.~{Van Gool}.
\newblock {Learning Driving Models with a Surround-View Camera System and a
  Route Planner}.
\newblock {\em ArXiv e-prints}, Mar. 2018.

\bibitem{lstm}
S.~Hochreiter and J.~Schmidhuber.
\newblock Long short-term memory.
\newblock {\em Neural computation}, 9(8):1735--1780, 1997.

\bibitem{car:knows:iccv15}
A.~Jain, H.~S. Koppula, B.~Raghavan, S.~Soh, and A.~Saxena.
\newblock Car that knows before you do: Anticipating maneuvers via learning
  temporal driving models.
\newblock In {\em IEEE International Conference on Computer Vision (ICCV)},
  2015.

\bibitem{sceneparse:opticalflow}
X.~Jin, H.~Xiao, X.~Shen, J.~Yang, Z.~Lin, Y.~Chen, Z.~Jie, J.~Feng, and
  S.~Yan.
\newblock Predicting scene parsing and motion dynamics in the future.
\newblock In {\em Advances in Neural Information Processing Systems (NIPS)}.
  2017.

\bibitem{intent:aware:pedestrian:prediction}
V.~Karasev, A.~Ayvaci, B.~Heisele, and S.~Soatto.
\newblock Intent-aware long-term prediction of pedestrian motion.
\newblock In {\em IEEE International Conference on Robotics and Automation
  (ICRA)}, 2016.

\bibitem{kasper2012object}
D.~Kasper, G.~Weidl, T.~Dang, G.~Breuel, A.~Tamke, A.~Wedel, and W.~Rosenstiel.
\newblock Object-oriented bayesian networks for detection of lane change
  maneuvers.
\newblock {\em IEEE Intelligent Transportation Systems Magazine}, 4(3):19--31,
  2012.

\bibitem{kendall2015bayesian}
A.~Kendall, V.~Badrinarayanan, and R.~Cipolla.
\newblock Bayesian segnet: Model uncertainty in deep convolutional
  encoder-decoder architectures for scene understanding.
\newblock {\em CoRR}, 2015.

\bibitem{driving:attention}
J.~Kim and J.~Canny.
\newblock Interpretable learning for self-driving cars by visualizing causal
  attention.
\newblock {\em arXiv:1703.10631}, 2017.

\bibitem{adam}
D.~Kingma and J.~Ba.
\newblock Adam: A method for stochastic optimization.
\newblock {\em arXiv:1412.6980}, 2014.

\bibitem{confidence:of:08}
C.~Kondermann, R.~Mester, and C.~Garbe.
\newblock A statistical confidence measure for optical flows.
\newblock In {\em European Conference on Computer Vision (ECCV)}. 2008.

\bibitem{completion:quality}
J.~Kopf, W.~Kienzle, S.~Drucker, and S.~B. Kang.
\newblock Quality prediction for image completion.
\newblock {\em ACM Trans. Graph.}, 31(6), 2012.

\bibitem{lecun2015deep}
Y.~LeCun, Y.~Bengio, and G.~Hinton.
\newblock Deep learning.
\newblock {\em Nature}, 521(7553):436--444, 2015.

\bibitem{LeCun:driving:05}
Y.~LeCun, U.~Muller, J.~Ben, E.~Cosatto, and B.~Flepp.
\newblock Off-road obstacle avoidance through end-to-end learning.
\newblock In {\em International Conference on Neural Information Processing
  Systems}, 2005.

\bibitem{predicting:deeper}
P.~Luc, N.~Neverova, C.~Couprie, J.~Verbeek, and Y.~LeCun.
\newblock Predicting deeper into the future of semantic segmentation.
\newblock In {\em International Conference on Computer Vision (ICCV)}, 2017.

\bibitem{oxford:driving}
W.~Maddern, G.~Pascoe, C.~Linegar, and P.~Newman.
\newblock 1 year, 1000 km: The oxford robotcar dataset.
\newblock {\em The International Journal of Robotics Research}, 36(1):3--15,
  2017.

\bibitem{looking:at:human}
E.~Ohn-Bar and M.~M. Trivedi.
\newblock Looking at humans in the age of self-driving and highly automated
  vehicles.
\newblock {\em IEEE Transactions on Intelligent Vehicles}, 1(1):90--104, 2016.

\bibitem{stereo:confidence}
M.-G. Park and K.-J. Yoon.
\newblock Leveraging stereo matching with learning-based confidence measures.
\newblock In {\em Computer Vision and Pattern Recognition (CVPR)}, 2015.

\bibitem{network:autonomous:1980}
D.~A. Pomerleau.
\newblock Advances in neural information processing systems 1.
\newblock chapter ALVINN: An Autonomous Land Vehicle in a Neural Network. 1989.

\bibitem{look:at:driver}
M.~Rezaei and R.~Klette.
\newblock Look at the driver, look at the road: No distraction! no accident!
\newblock In {\em IEEE Conference on Computer Vision and Pattern Recognition},
  2014.

\bibitem{semantic:foggy:scene}
C.~{Sakaridis}, D.~{Dai}, and L.~{Van Gool}.
\newblock {Semantic Foggy Scene Understanding with Synthetic Data}.
\newblock {\em International Journal of Computer Vision (IJCV)}, 2018.

\bibitem{control:driver:modeling}
V.~A. Shia, Y.~Gao, R.~Vasudevan, K.~D. Campbell, T.~Lin, F.~Borrelli, and
  R.~Bajcsy.
\newblock Semiautonomous vehicular control using driver modeling.
\newblock {\em IEEE Transactions on Intelligent Transportation Systems},
  15(6):2696--2709, 2014.

\bibitem{vgg16}
K.~Simonyan and A.~Zisserman.
\newblock Very deep convolutional networks for large-scale image recognition.
\newblock In {\em ICLR}, 2015.

\bibitem{driveability:maps:for:lane}
S.~Sivaraman and M.~M. Trivedi.
\newblock Dynamic probabilistic drivability maps for lane change and merge
  driver assistance.
\newblock {\em IEEE Transactions on Intelligent Transportation Systems},
  15(5):2063--2073, Oct 2014.

\bibitem{Anticipating:accidents:DB:18}
T.~{Suzuki}, H.~{Kataoka}, Y.~{Aoki}, and Y.~{Satoh}.
\newblock Anticipating traffic accidents with adaptive loss and large-scale
  incident db.
\newblock {\em ArXiv e-prints}, 2018.

\bibitem{gaze:driver}
F.~Vicente, Z.~Huang, X.~Xiong, F.~De~la Torre, W.~Zhang, and D.~Levi.
\newblock Driver gaze tracking and eyes off the road detection system.
\newblock {\em IEEE Transactions on Intelligent Transportation Systems},
  16(4):2014--2027, 2015.

\bibitem{end:driving:16}
H.~Xu, Y.~Gao, F.~Yu, and T.~Darrell.
\newblock End-to-end learning of driving models from large-scale video
  datasets.
\newblock In {\em IEEE Computer Vision and Pattern Recognition (CVPR)}, 2017.

\bibitem{zhang2014predicting}
P.~Zhang, J.~Wang, A.~Farhadi, M.~Hebert, and D.~Parikh.
\newblock Predicting failures of vision systems.
\newblock In {\em IEEE Conference on Computer Vision and Pattern Recognition
  (CVPR)}, 2014.

\end{thebibliography}
\end{document}